\documentclass{article}
\usepackage{ijcai20}

\usepackage{times}

\usepackage{soul}
\usepackage{url}
\usepackage[hidelinks]{hyperref}
\usepackage[utf8]{inputenc}
\usepackage[small]{caption}
\usepackage{graphicx}
\usepackage{amsmath}
\usepackage{booktabs}
\usepackage{algorithm}
\usepackage{algorithmic}
\usepackage{subfigure}
\usepackage{stfloats}

\title{A Strong Baseline for Crowd Counting and Unsupervised People Localization}

\author{
Liangzi Rong$^1$\And
Chunping Li$^1$\footnote{Contact Author}\\
\affiliations
$^1$School of Software, Tsinghua University, Beijing, China\\
\emails
rlz17@mails.tsinghua.edu.cn,
cli@tsinghua.edu.cn
}

\begin{document}

\maketitle

\begin{abstract}
In this paper, we explore a strong baseline for crowd counting and an unsupervised people localization algorithm based on estimated density maps. Firstly, existing methods achieve state-of-the-art performance based on different backbones and kinds of training tricks. We collect different backbones and training tricks and evaluate the impact of changing them and develop an efficient pipeline for crowd counting, which decreases MAE and RMSE significantly on multiple datasets. We also propose a clustering algorithm named isolated KMeans to locate the heads in density maps. This method can divide the density maps into subregions and find the centers under local count constraints without training any parameter and can be integrated with existing methods easily.
\end{abstract}

\section{Introduction}

Crowd counting is a computer vision task which aims to output a density map indicating the distribution of crowd and get the estimated count of people by calculating the integral over the map in still images. In recent years, many novel networks are designed and the MAE on ShanghaiTechA dataset has been improved from 110.2(MCNN) to 61.7(SPN). It is important to note, however, that the improvement is not only the result of newly proposed methods. It also benefits from various training tricks in implementation. According to \cite{he2019bag} training details including data preprocessing, changes of loss function, the ground truth generation, and learning objectives setting make a big difference.

\begin{figure}[t]
    \centering
    \subfigure[Overview of our strong baseline for crowd counting and people localization.]{ 
    \includegraphics[width=0.95\columnwidth]{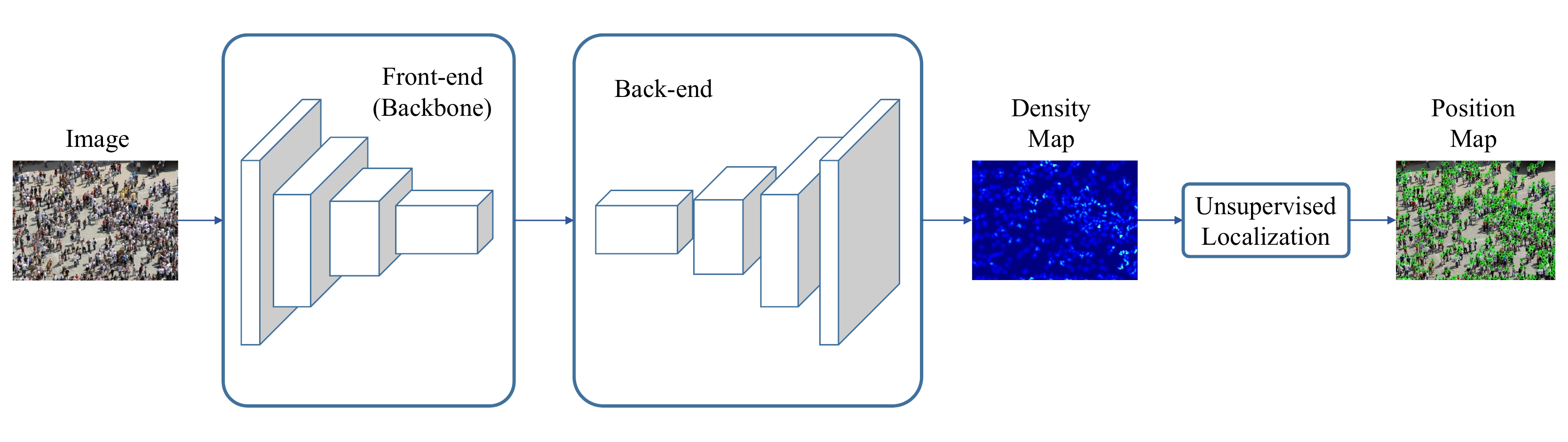}
    \label{fig:pipeline}}
    \subfigure[Comparison of MAE and RMSE curve on ShanghaiTech PartA dataset.]{\includegraphics[width=0.95\columnwidth]{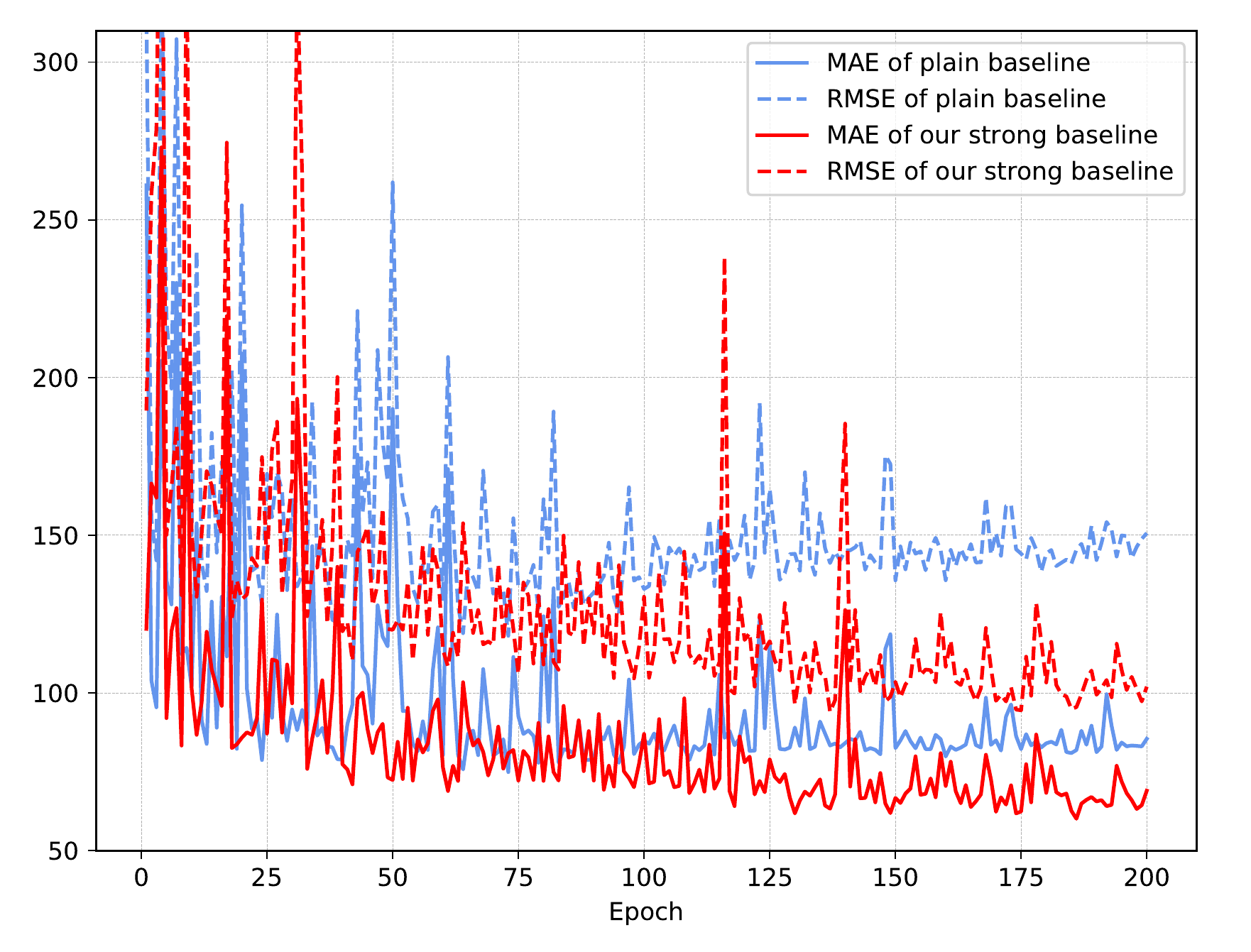}
    \label{fig:mae_curve}}
    \caption{Overview of our strong baseline and results.}
    \label{fig:overview}
\end{figure}

Since \cite{li2018csrnet}, most researchers extend their work using VGG-16 as the backbone, and the main components of the fully convolutional networks are like the front-end and back-end in \ref{fig:pipeline}. Existing method \cite{wang2019segmentation} has shown that training tricks can make a big difference. Besides, there are also some methods using other classification networks as the backbone. To make a fair comparison and construct a strong baseline easy to follow for the academia and the industry, we collect and evaluate a set of backbones and tricks for regressing a more accurate density map. As Figure \ref{fig:mae_curve} shows, by combining these tricks, the MAE can be decreased from 74.9 to 60.1, and RMSE can be decreased from 126.1 to 95.5, which is better than SPN. They show consistent improvement on multiple datasets which proves the effectiveness of this strong baseline.

Another drawback of existing previous methods is the lack of position information for each person. Note that position information is essential for us to understanding the crowd images, we extend our work to find the position of each person. \cite{liu2019recurrent} and \cite{liu2019point} have combined detection networks to located heads. However, there are no standard labels in crowd counting dataset as in detection datasets, so the ground truth is not reliable. Besides, we have to retrain the parameters of detection networks for different datasets which tend to be time consuming in practical use. In fact, the estimated density maps have indicated the positions by showing clusters around each head center. A human can find most of the heads easily by observing the clusters. Based on this observation, we proposed an unsupervised localization method named isolated KMeans based on cluster algorithms. Because people count $n$ can be obtained by calculating the integral, the objective of this problem is simplified into finding $n$ centers indicating the position of $n$ persons. Each point whose value is bigger than 0 in density maps is regarded as a potential head and cluster centers converge to the points near the heads. In addition, isolated KMeans can locate the head centers under local count constraints to make the number of cluster centers and people match both globally and locally. We can integrate this localization method with any other existing crowd counting methods without training numerous parameters. To the best of our knowledge, we are the first to use unsupervised method to do people localization in this field.

To summarize, our main contribution is three-fold:

\begin{itemize}
    \item We collect a series of backbones and training tricks for crowd counting and evaluate their impacts on the model performance to validate the effectiveness and make a fair comparison.
    \item We develop a strong baseline by combining the effective tricks which can improve the accuracy and density map quality significantly on three widely used datasets.
    \item We propose an unsupervised method for people localization in density maps. It is achieved using a cluster algorithm named isolated KMeans which divides one density map into subregions and locate the heads under local people count constraints. This method frees us of training a large number of parameters of detection networks and can be integrated with any existing counting method.
\end{itemize}

\section{The Setup of the Standard Baseline}
All datasets only provide discrete point labels in the centers of heads, which is hard to learn. We firstly convert the labeled points into density maps with continuous distribution.

\textbf{Ground Truth Density Map Generation.} Following most previous works, each labeled point at a head's center is substituted with a Gaussian distribution, and superimposing multiple Gaussian distributions produces the ground truth density map. This is formulated as 
\begin{equation}
    M_{i}=\sum_{j=1}^{n}\delta(x-head_{j})*G_{\sigma _{j}}(x)\label{dm_gen}
\end{equation}
where $M_{i}$ means the density map, and $head_{j}$ is the coordinates of $j_{th}$ head annotation. $G(x)$ is the Gaussian distribution. $n$ is the number of head annotations. In ShanghaiTech PartB dataset, the standard deviation parameter $\sigma_{j}$ is set to 15. In other datasets, $\sigma_{j}$ is calculated as $\sigma_{j} = \beta*d_{j}$ where $d_{j}$ is the average distance of $x_{j}$ and its $k$ nearest neighbors. In this paper we adopt the configuration of $\beta=0.3$ and $k=3$.

\textbf{Training Settings and Evaluation Metrics.} Experiments are conducted on ShanghaiTech PartA dataset for comparison. We use PyTorch to implement all training process. For all models, the total training epochs is set to 200 and initial learning rate is set to 1e-5. Adam optimizer and ReduceOnPlateau learning rate scheduler are used to optimize the parameters. We adopt two mainly used metrics MAE and RMSE for evaluating the accuracy of counting and the quality of the density maps.

\textbf{Backbones.} Most existing methods adopt VGG-16 as the backbone. Some literature also adopts ResNet-50\cite{wang2019learning} and InceptionV3\cite{wang2019segmentation}. Similarly to \cite{li2018csrnet}, we remove the fully connected layers and preserve the convolutional layers to compare the performance of these backbones and to explore if batch normalization should be used or not. For back-end, we use the configuration of $3 \times CS(512,3,2) - Upsample - CS(256,3,2) - Upsample - CS(128,3,2) - Upsample - CS(64,3,2) - Upsample - C(1,1,1)$. $N \times CS(m,s,d)$ means N convolutional layers integrated with SE block \cite{hu2018squeeze} with $m$ filters whose size is $s \times s$ and dilation rate is $d$ followed by the Swish activation layer by \cite{ramachandran2017swish}. We use bilinear interpolation to upsample the feature maps.

\begin{table}[htbp]
\centering
\begin{tabular}{lrrrr}  
\toprule
Backbones  & MAE & RMSE\\
\midrule
first 10 layers of VGG-16    & \textbf{74.8}  & 126.1 \\
first 13 layers of VGG-16    & 74.9  & \textbf{118.1}  \\
\midrule
first 13 layers of VGG-16-bn    & 92.1  & 134.6 \\
InceptionV3    & 119.4  & 170.5 \\
ResNet-50   & 80.6  & 130.1 \\

\bottomrule
\end{tabular}
\caption{Performance of different backbones.}
\label{tab:backbones}
\end{table}

For models without batch normalization, we train them using original images of different sizes. For models with batch normalization, we train them using $256 \times 256$ patches randomly cropped from the images. To be specific, we randomly choose 4 images at a time and 4 patches are randomly cropped from each of them to constitute a batch. Finally one batch consists of 16 patches of $256 \times 256$. In experiments, if the back-end contains batch normalization layers, the model converge to output 0. So batch normalization layers only exist in front-end. The performance of different backbones is reported in Table \ref{tab:backbones}. Due to the superior performance, we select first 13 layers of VGG-16 as the backbone and conduct following experiments.

\section{Loss Functions}
Many loss functions are proposed to improve the quality of density maps. Among them, \textbf{MSE loss ($\ell_{MSE}$)} is the mostly used one for its simplicity and effectiveness. It is defined as:
\begin{equation}
    \ell_{MSE} = \frac{1}{N}\sum_{i=1}^{N}||\hat{M_{i}}-M_{i}||_{2}
\end{equation}

\textbf{Spatial Abstraction Loss ($SAL$)} is proposed by \cite{jiang2019crowd} which progressively computes the MSE losses on multiple abstraction levels. The computation is formalized as:
\begin{equation}
    SAL = \sum_{k_{a}=1}^{K_{a}}\frac{1}{N_{k_{a}}}||\hat{M_{i}^{k_{a}}}-M_{i}^{k_{a}}||_{2}
\end{equation}
where $k_{a}$ indicates the abstraction level and $K_{a}$ is set to 3. $\hat{M_{i}^{k_{a}}}$ and $M_{i}^{k_{a}}$ are the downsampled density maps by pooling layers and $N_{k_{a}}$ is the number of pixels within a map.

\textbf{Multi-scale density level consistency loss ($\ell_{c}$)} is proposed by \cite{dai2019dense}. $\ell_{c}$ loss separates the density maps into subregions using adaptive average pooling and calculates L1 loss to enforce the consistency at different scale levels. It is defined as:
\begin{equation}
    \ell_{c} = \sum_{s=1}^{S}\frac{1}{k_{s}^{2}}||\hat{M_{i}^{s}}-M_{i}^{s}||_{1}
\end{equation}
where $s$ is the scale levels, and $k_{s}$ is size of density maps. They are set to 3 and $1 \times 1, 2 \times 2, 4 \times 4$ respectively.

SSIM is proposed to measure the similarity of images by\cite{wang2004image} and \textbf{Local Pattern Consistency Loss ($\ell_{SSIM}$)} is introduced by \cite{cao2018scale} to enhance the structure consistency. A fixed-parameter kernel $W(p)$ to adopted to define the weights of different positions in one sliding window. For the same location $x$ in density maps $\hat{M_{i}}$ and $M_{i}$, the local statistics are defined as:
\begin{equation}
    \mu_{\hat{M_{i}}}({x})=\sum_{p \in \mathcal{P}} W(p) \cdot \hat{M_{i}}(x+p)
\end{equation}
\begin{equation}
    \sigma_{\hat{M_{i}}}^{2}(x)=\sum_{p \in \mathcal{P}} W(p) \cdot\left[\hat{M_{i}}(x+p)-\mu_{\hat{M_{i}}}(x)\right]^{2}
\end{equation}
\begin{equation}
\begin{aligned}
    \sigma_{\hat{M_{i}}\,M_{i}}(x)&=\sum_{p \in \mathcal{P}} W(p)
    \cdot\left[\hat{M_{i}}(x+p)-\mu_{\hat{M_{i}}}(x)\right]\\ &\cdot\left[M_{i}(x+p)-\mu_{M_{i}}(x)\right]
    \end{aligned}
\end{equation}
and the SSIM index and SSIM loss are calculated as:
\begin{equation}
    SSIM=\frac{\left(2 \mu_{\hat{M_{i}}} \mu_{M_{i}}+C_{1}\right)\left(2 \sigma_{\hat{M_{i}}\, M_{i}}+C_{2}\right)}{\left(\mu_{\hat{M_{i}}}^{2}+\mu_{M_{i}}^{2}+C_{1}\right)\left(\sigma_{\hat{M_{i}}}^{2}+\sigma_{M_{i}}^{2}+C_{2}\right)}
\end{equation}
\begin{equation}
    \ell_{SSIM} = 1 - SSIM
\end{equation}

We train the model using different loss functions, and the results are reported in Table \ref{tab:loss}.

\begin{table}[ht]
    \centering
    \begin{tabular}{lrr}
    \toprule
    Loss Functions&MAE&RMSE\\
    \midrule
    $\ell_{MSE}$  &74.9 &118.1  \\
    $\ell_{MSE}$ + $SAL(Max\, Pooling)$ & 149.0&225.4\\
    $\ell_{MSE}$ + $SAL(Avg\, Pooling)$&128.4&200.9\\
    $\ell_{MSE} + \ell_{c}$ &\textbf{67.8}&105.7\\
    $\ell_{MSE} + \ell_{SSIM}$&90.1&134.6\\
    $\ell_{SSIM}$ &71.5&115.0\\
    
    \bottomrule
    \end{tabular}
    \caption{Performance of different loss functions.}
    \label{tab:loss}
\end{table}
As we can see, $\ell_{MSE} + \ell_{c}$ gets the best MAE performance. Also because of its simple form, we select it as the loss function.

\section{Training Tricks}
We collect various tricks in training and evaluate each of them. By combining the effective tricks, we can boost the model's accuracy without changing the architecture.

\textbf{Data Augmentation.} In crowd counting, estimated density maps are sensitive to the size of people. Resize and rotation is harmful to the performance of the network, so we apply cropping to enlarge the training set. There are 4 kinds of cropping: \textbf{(\romannumeral1) random 0.3 - random0.9} means randomly cropping by ratio 0.3 - 0.9 from each image. \textbf{(\romannumeral2) fixed 0.5} means cropping 4 non-overlapping quarters from fixed locations in original images. \textbf{(\romannumeral3) fixed + random 0.5} means 4 patches are cropped from fixed locations and 5 patches randomly cropped from each image. \textbf{(\romannumeral4) mixed} means randomly cropping by a random ratio in \{0.3, 0.4, 0.5, 0.6, 0.7\}. Table \ref{tab:tricks} shows that random 0.3 is the most effective data augmentation. In addition, cropping can also reduce the training time.

\textbf{Curriculum learning.} In \cite{wang2019segmentation}, the curriculum is designed based on the fact that dense crowds are more difficult to count than sparse crowd. In training process, the weights of areas with higher density are relative lower and gradually increased to equal to areas with lower density. This is implemented by a weight matrix $W(e)$ which is defined as:
\begin{equation}
    W(e)=\frac{T(e)}{\max \left\{M_{i}, T(e)\right\}}
\end{equation}

$T(e)$ is a threshold matrix which is defined as:
\begin{equation}
    T(e)=k_{e} e+b_{e}
\end{equation}
where $k_{e}$ and $b_{e}$ are coefficients determined by prior knowledge and $e$ denotes the epoch number. $W$ has the same size as the density map $M_{i}$. Then $\ell_{MSE}$ is calculated as:
\begin{equation}
    \ell_{MSE}= \frac{1}{N}\sum_{i=1}^{N}||W(e)\odot(\hat{M_{i}}-M_{i})||_{2}
\end{equation}
Empirically, we set $k_{e}=2e-3, b_{e}=5e-3$. Using curriculum learning can decrease MAE to 63.3 as Table \ref{tab:tricks} shows.

\begin{figure}[htbp]
    \centering
    \includegraphics[width=0.95\columnwidth]{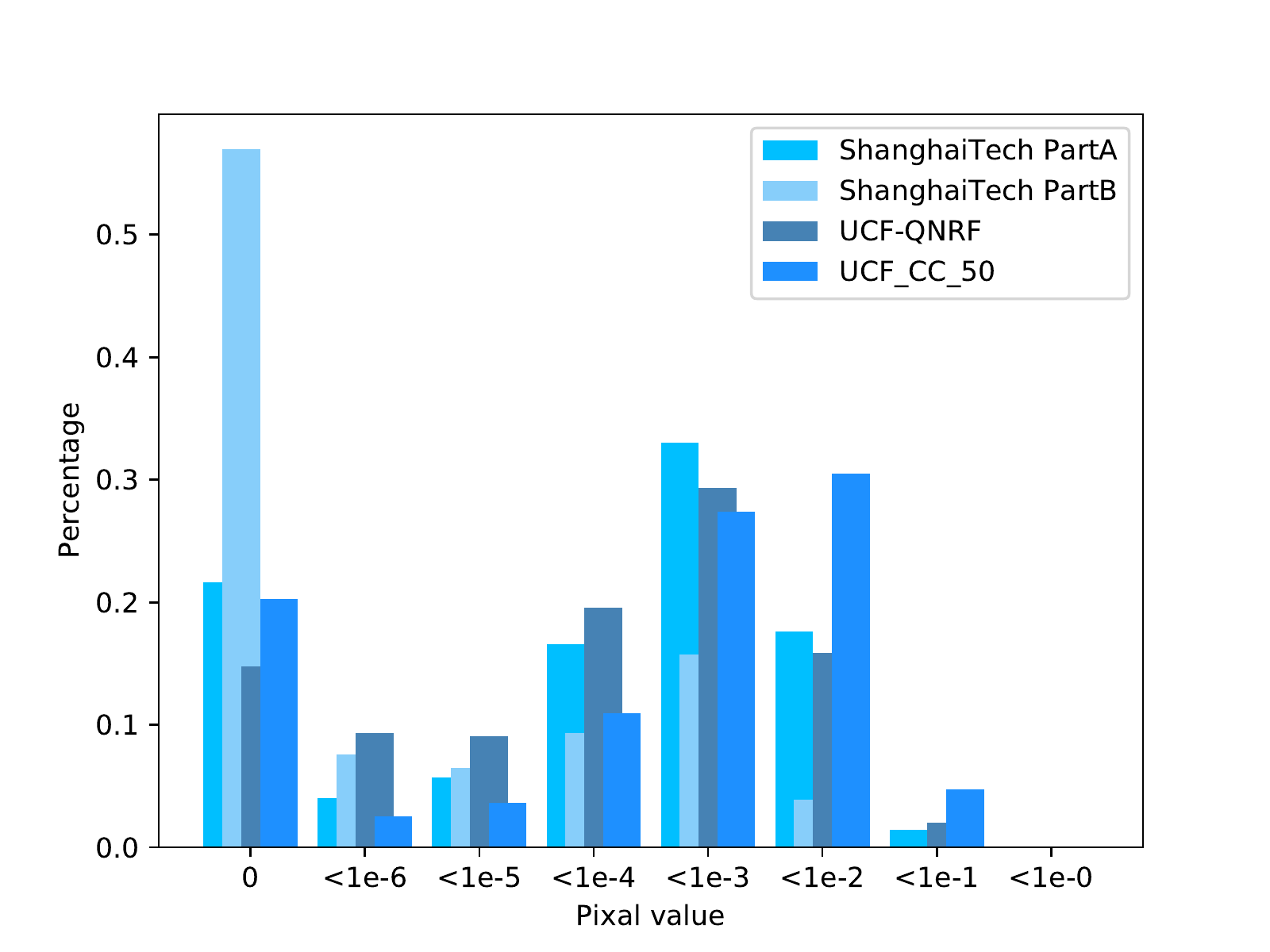}
    \caption{Statistics of pixel values on three datasets.}
    \label{fig:value_stats}
\end{figure}

\textbf{Value Expansion.} The statistics of pixel values in mainly used datasets can be viewed in Figure \ref{fig:value_stats}. Many pixel values are very small which can lead to the loss of precision.

To facilitate the training process, we multiply the ground truth density map by a scale factor $k_{expand}$. In inference process, we can multiply the estimated density map by $\frac{1}{k_{expand}}$. Using value expansion, we can improve the MAE to 62.6.

\textbf{Validate by Patch.} \cite{cao2018scale} firstly proposes the patch-based validation strategy. In validation process, each image is divided into several quarters, and quarters are fed into the network. By calculating the sum of each quarter's output, we can get the overall count. Table \ref{tab:tricks} reports that RMSE can be improved to 95.4 using this strategy.

\begin{table}[ht]
    \centering
    \begin{tabular}{lrr}
    \toprule
    Training Tricks&MAE&RMSE\\
    \midrule
    Baseline &67.8&105.7\\
    random 0.3&\textbf{64.9}&100.5\\
    random 0.4&66.3&99.4\\
    random 0.5&66.0&\textbf{97.8}\\
    random 0.6&66.0&98.9\\
    random 0.7&66.5&98.1\\
    random 0.8&67.9&100.7\\
    random 0.9&68.0&100.9\\
    fixed 0.5&68.7&104.0\\
    fixed + random 0.5 &65.6&100.8\\
    mixed &69.0&103.6\\
    \midrule
    + curriculum learning &63.3 &97.3\\
    + value expansion 10&\textbf{62.6}&97.8\\
    + value expansion 100&66.4&101.6\\
    + value expansion 1000&67.7&105.1\\
    + validate by patch&63.1&\textbf{95.4}\\
    \bottomrule
    \end{tabular}
    \caption{Performance of adding training tricks.}
    \label{tab:tricks}
\end{table}

\section{Learning Objectives}
 
\begin{figure}[htbp]
    \centering
    \includegraphics[width=\columnwidth]{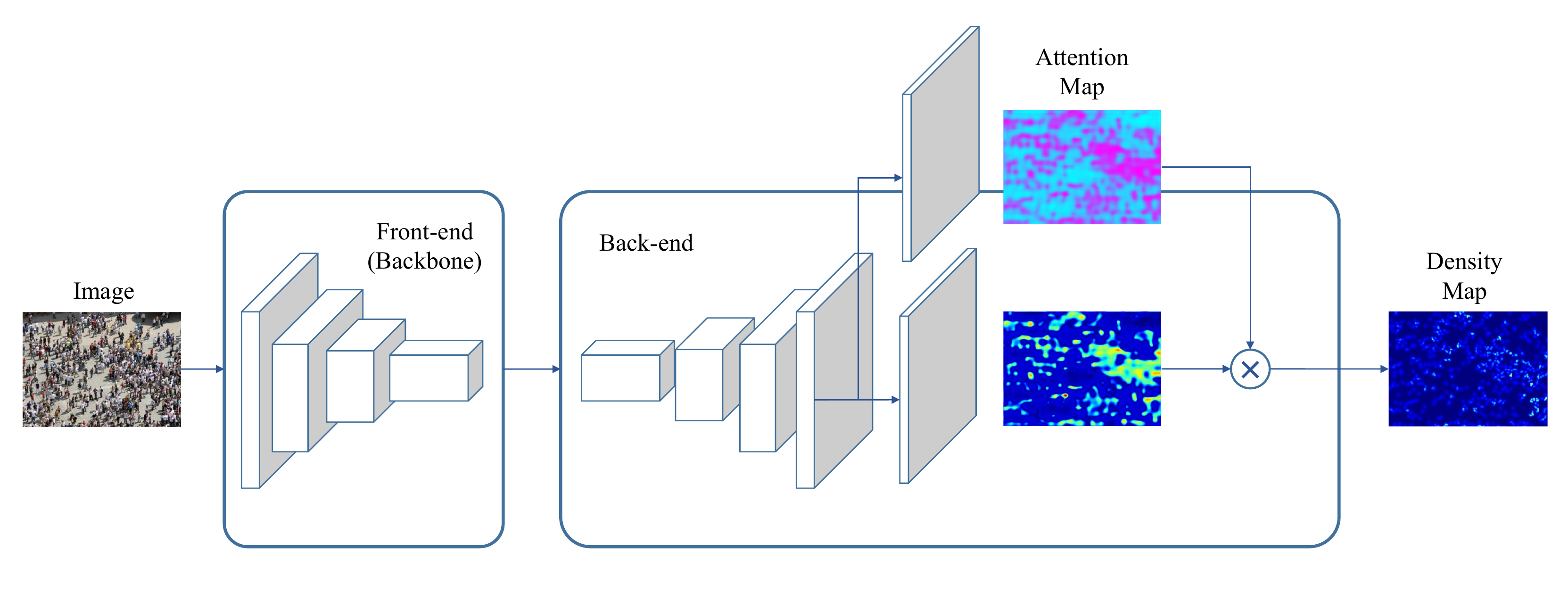}
    \caption{Network with an attention branch inserted.}
    \label{fig:attention_network}
\end{figure}

Besides density maps, \cite{liu2019adcrowdnet} and \cite{wang2019segmentation} propose using attention maps to emphasize the crowd regions and weaken the impact of background regions. An extra branch with the configuration of $2 \times CS(64,3,2) - C(1,1,1)- Sigmoid$ is inserted after the penultimate layer and can produce a one-channel attention map $M^{att}_{i}$ with same size of density map $M_{i}$. This branch aims to regress an attention map whose value is close to 1 in foreground regions and close to 0 in background region and is illustrated in Figure \ref{fig:attention_network}.

\textbf{Ground Truth Attention Map Generation.} There are two ways of generating attention maps. \textbf{(\romannumeral1) Window-based.} We set a $25 \times 25$ windows centered at each labeled point. Values within those windows are set to 1 and those out of them are set to 0. \textbf{(\romannumeral2) Threshold-based.} We set the threshold at the value bigger than 40\% of the values. We test the two ways, and way (\romannumeral1) shows a more accurate result.

\textbf{Attention Map Loss.} The loss function of attention map is defined as 
\begin{equation}
    \begin{aligned}
\ell^{att}(\Theta) &= \| M_{i}^{att} \odot \log \left(\hat{M}_{i}^{att}\right) \\
&+\left(1-M_{i}^{att}\right) \odot \log \left(1-\hat{M}_{i}^{att}\right) \|_{1}
\end{aligned}
\end{equation}
The total loss is a weighted sum:$\mathcal{L}_{total}=\ell_{MSE}+\ell_{C}+\lambda\ell^{att}$. Empirically, when $\lambda=0.5$, the model gets best performance.

\begin{figure*}[ht]
     \centering
     \subfigure[Image]{
     \begin{minipage}[b]{0.38\columnwidth} 
      \includegraphics[width=\columnwidth]{dm/IMG_175.jpg}\\
      \includegraphics[width=\columnwidth]{dm/IMG_78.jpg}\
        \end{minipage}
    }
    \subfigure[Baseline]{
     \begin{minipage}[b]{0.38\columnwidth} 
      \includegraphics[width=\columnwidth]{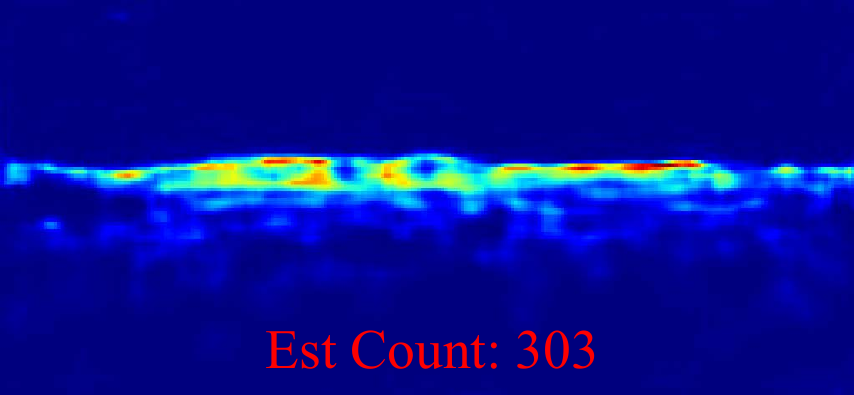}\\
      \includegraphics[width=\columnwidth]{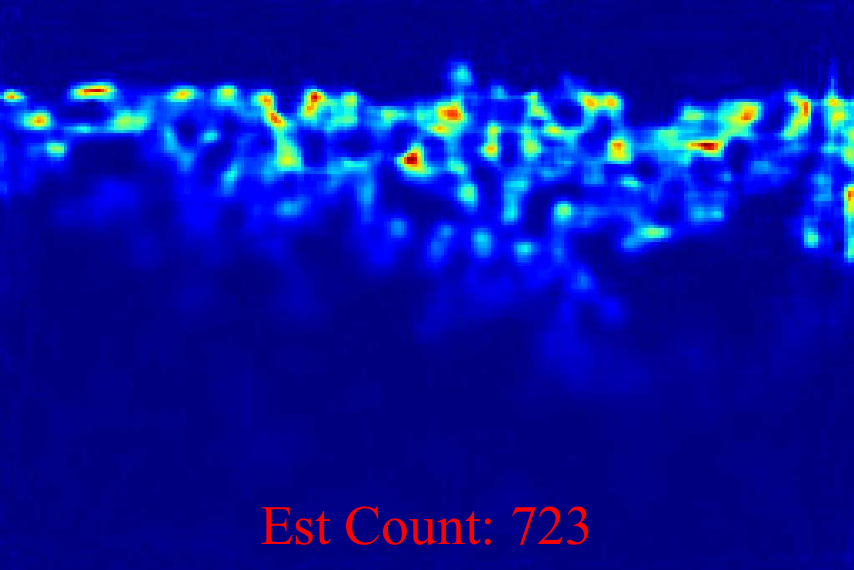}\
        \end{minipage}
    }
     \subfigure[Our strong baseline]{
     \begin{minipage}[b]{0.38\columnwidth} 
      \includegraphics[width=\columnwidth]{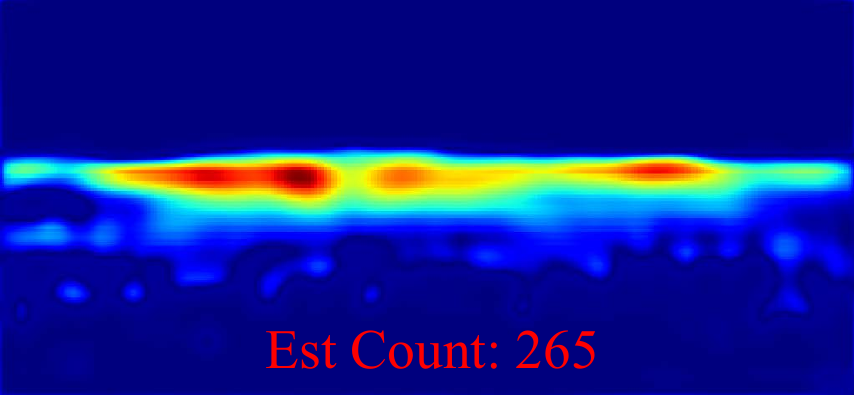}\\
      \includegraphics[width=\columnwidth]{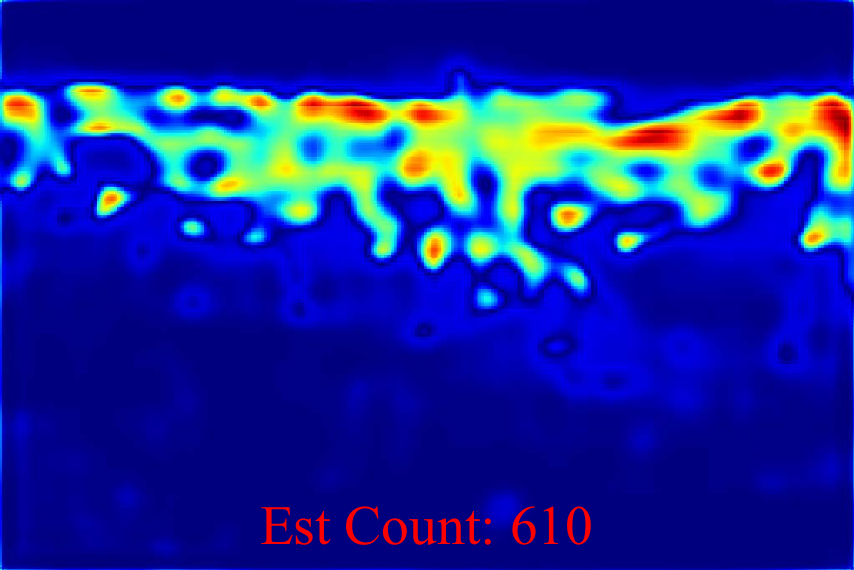}\
        \end{minipage}
    }
    \subfigure[Attention maps]{
     \begin{minipage}[b]{0.38\columnwidth} 
      \includegraphics[width=\columnwidth]{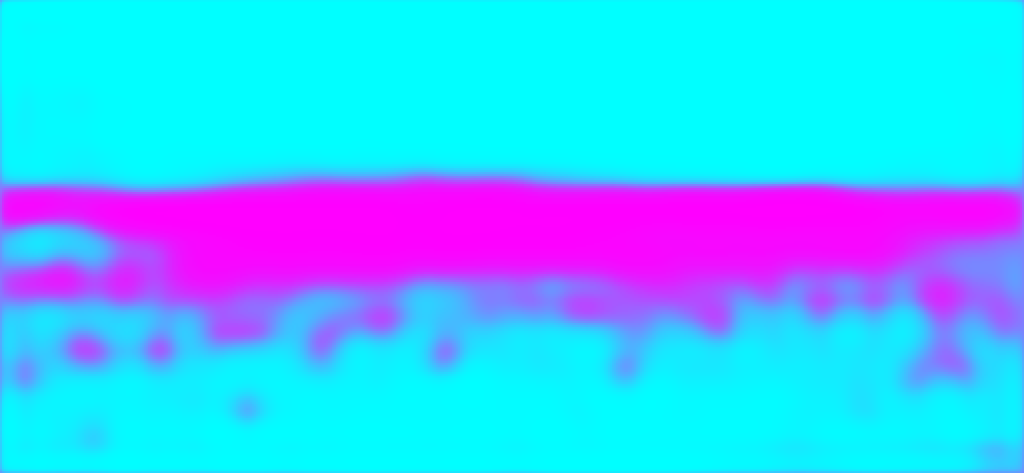}\\
      \includegraphics[width=\columnwidth]{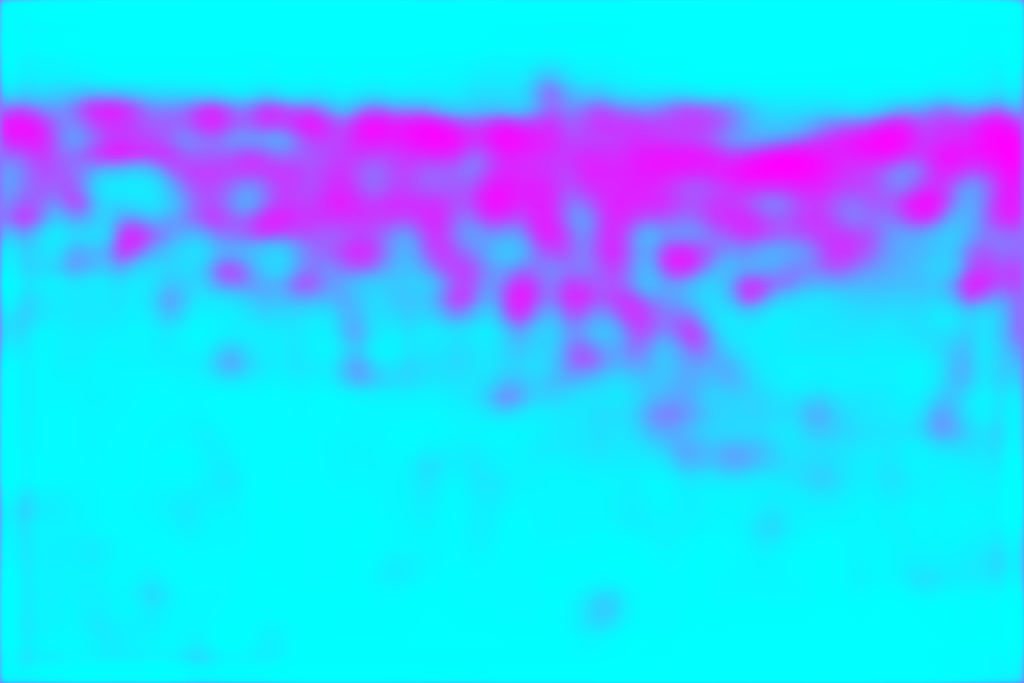}\
        \end{minipage}
    }
    \subfigure[Ground truth]{
     \begin{minipage}[b]{0.38\columnwidth} 
      \includegraphics[width=\columnwidth]{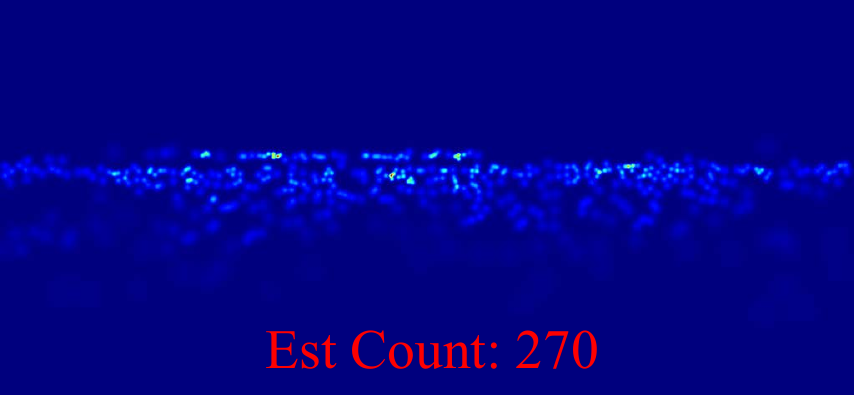}\\
      \includegraphics[width=\columnwidth]{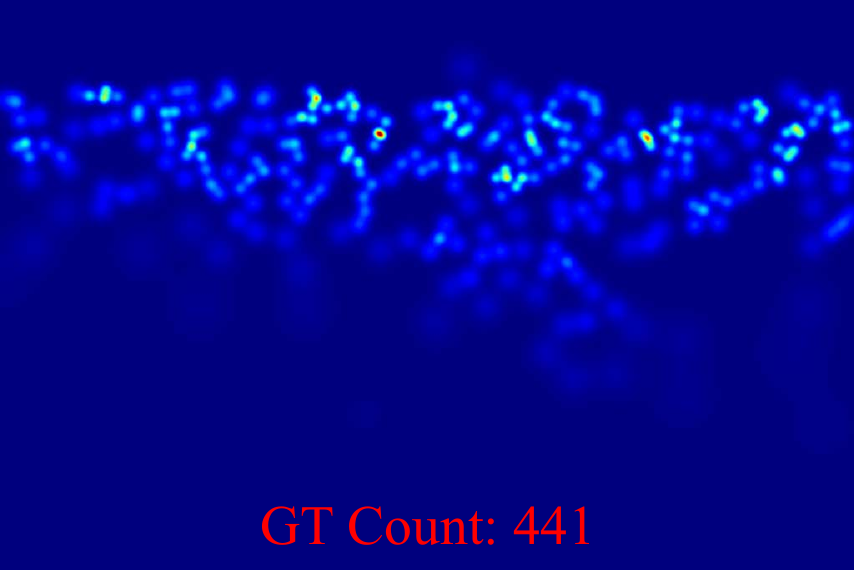}\
        \end{minipage}
    }
     \caption{Visualization of estimated density maps and attention maps on ShanghaiTech dataset.}
     \label{fig:density_maps}
 \end{figure*}
 
\textbf{Density Map Size.} We change the number of upsample layers to evaluate the impact of different density map sizes. By adding 1 - 4 upsample layers, we can get $\frac{1}{8}\times\frac{1}{8}$ - $1\times1$ size of density maps. Table \ref{tab:obj} shows that estimating the same size of density maps benefits the performance. 
\begin{table}[ht]
    \centering
    \begin{tabular}{lrr}
    \toprule
    Learning Objectives&MAE&RMSE\\
    \midrule
    Baseline&63.1&95.4\\
    +AM(Window-based) $\lambda=0.02$&66.8&99.3\\
    +AM(Threshold-based) $\lambda=0.02$&65.8&100.6\\
    +AM $\lambda=0.02$&61.8&96.9\\
    +AM $\lambda=0.1$&65.8&100.6\\
    +AM $\lambda=0.5$&\textbf{60.1}&\textbf{95.5}\\
    \midrule
    $\frac{1}{8}\times\frac{1}{8}$ size&67.2&101.8\\
    $\frac{1}{4}\times\frac{1}{4}$ size&64.7&101.4\\
    $\frac{1}{2}\times\frac{1}{2}$ size&63.8&\textbf{95.0}\\
    $1\times1$ size&\textbf{60.1}&95.5\\
    \bottomrule
    \end{tabular}
    \caption{Performance of different learning objectives. AM means attention map.}
    \label{tab:obj}
\end{table}

\section{Comparisons with State-of-the-art}

We evaluate our final model with other state-of-the-art methods on three mainly used datasets.

\textbf{ShanghaiTech} dataset is introduced by \cite{zhang2016single} which consists of PartA and PartB. PartA has 300 training images and 182 testing images with relatively high density. PartB has 400 training images and 316 testing images with relatively low density.As table \ref{tab:sht} shows, our strong baseline decreases MAE by 2.6\% on PartA and RMSE by 0.8\% on PartB.
\begin{table}[ht]
    \centering
    \begin{tabular}{lrrrr}
    \toprule
    &\multicolumn{2}{c}{PartA}&\multicolumn{2}{c}{PartB}\\
    
    Models&MAE&RMSE&MAE&RMSE\\
    \midrule
    \cite{li2018csrnet}&68.2&115.0&10.6&16.0\\
    \cite{cao2018scale}&67.0&104.5&8.4&13.6\\
    \cite{jiang2019crowd}&64.2&109.1&8.2&12.8\\
    \cite{liu2019adcrowdnet}&63.2&98.9&7.7&12.9\\
    \cite{liu2019context}&62.3&100.0&7.8&12.2\\
    \cite{liu2019recurrent}&65.1&106.7&8.4&14.1\\
    \cite{shi2019revisiting}&62.4&102.0&\textbf{7.6}&11.8\\
    \cite{wang2019learning}&64.8&107.5&7.6&13.0\\
    \cite{chen2019scale}&61.7&99.5&9.4&14.4\\
    \cite{wan2019residual}&63.1&96.2&8.7&13.6\\
    Ours&\textbf{60.1}&\textbf{95.5}&7.9&\textbf{11.7}\\
    \bottomrule
    \end{tabular}
    \caption{Comparisons on ShanghaiTech dataset.}
    \label{tab:sht}
\end{table}

\textbf{UCF-QNRF} dataset is introduced by \cite{idrees2018composition}. It is has 1201 training images and 334 testing images with high resolution. Performance on this dataset is reported in Table \ref{tab:ucf}, and our strong baseline decreases MAE by 2.7\%.

\textbf{UCF\_CC\_50} dataset is introduced by \cite{idrees2013multi} including 50 crowd images with extremely high density. We follow the standard 5-fold cross-validation. It can be viewed in Table \ref{tab:ucf} that our baseline decreases the MAE by 4.0\%.

\begin{table}[!hbtp]
    \centering
    \begin{tabular}{lrrrr}
    \toprule
    &\multicolumn{2}{c}{UCF-QNRF}&\multicolumn{2}{c}{UCF\_CC\_50}\\
    Models&MAE&RMSE&MAE&RMSE\\
    \midrule
    \cite{li2018csrnet}&-&-&266.1&397.5\\
    \cite{cao2018scale}&-&-&258.4&334.9\\
    \cite{jiang2019crowd}&113.0&188.0&249.4&354.5\\
    \cite{liu2019adcrowdnet}&-&-&257.1&363.5\\
    \cite{liu2019context}&107.0&183.0&212.2&\textbf{243.7}\\
    \cite{liu2019recurrent}&116.0&195.0&-&-\\
    \cite{shi2019revisiting}&-&-&241.7&320.7\\
    \cite{wang2019learning}&102.0&\textbf{171.4}&214.2&318.2\\
    \cite{chen2019scale}&-&-&259.2&335.9\\
    \cite{wan2019residual}&-&-&355.0&560.2\\
    Ours&\textbf{99.2}&179.1&\textbf{203.7}&283.1\\
    \bottomrule
    \end{tabular}
    \caption{Comparisons on UCF-QNRF and UCF\_CC\_50 dataset.}
    \label{tab:ucf}
\end{table}

\section{Unsupervised People Localization}

\subsection{Point Set Construction}
\begin{figure}[htbp]
    \centering
    \includegraphics[width=0.9\columnwidth]{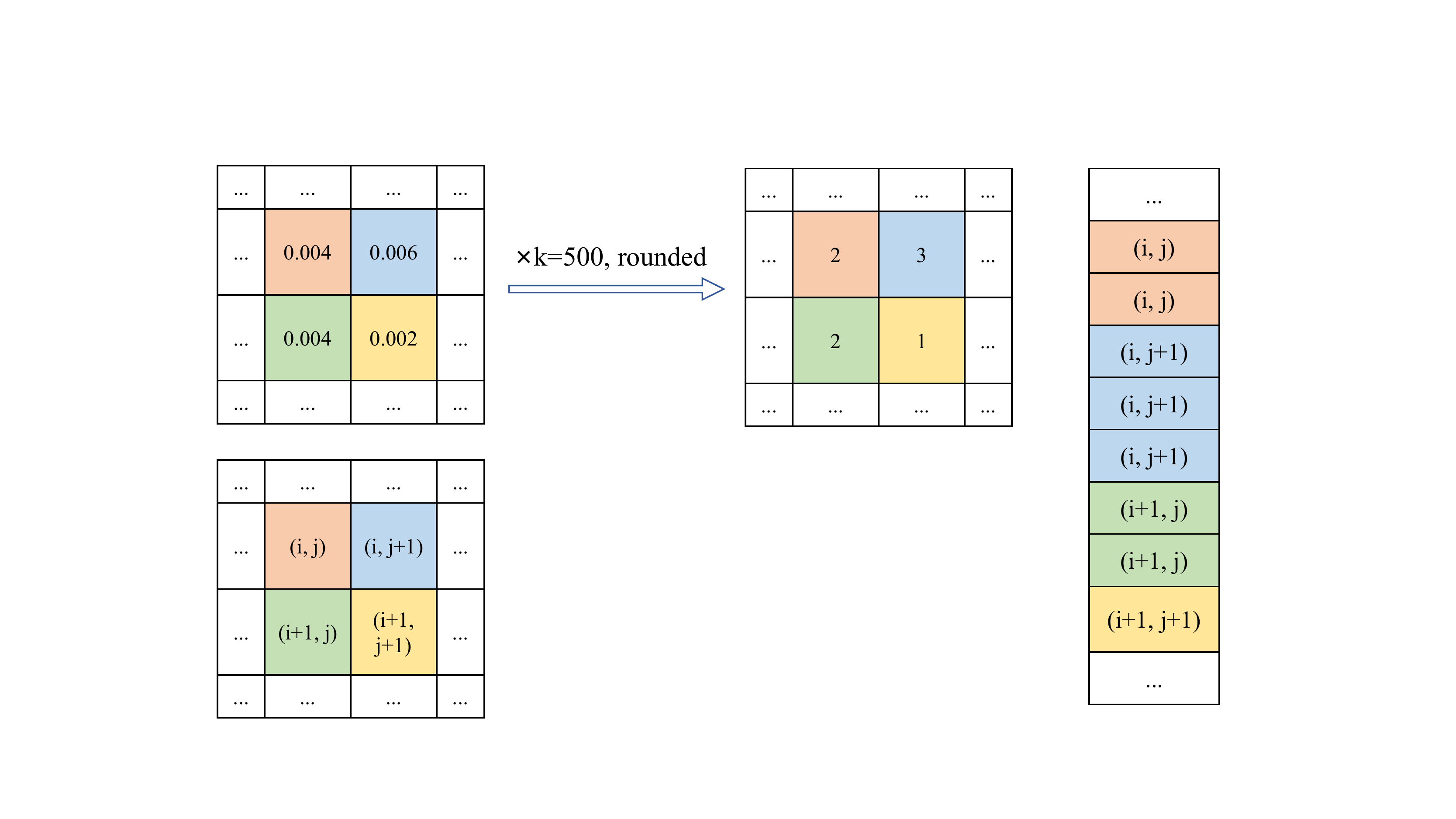}
    \caption{Point set construction.}
    \label{fig:data_construct}
\end{figure}
Firstly, we use the density maps to construct a point set. Density maps are multiplied by an expansion factor $k$ and rounded, so each pixel has an integer to indicate its frequency. In experiments, $k$ is set to 500. We sample the coordinate $(i,j)$ of each pixel by its frequency as Figure \ref{fig:data_construct} shows and construct a point set.

\subsection{KMeans}
\begin{figure*}[htbp]
     \centering
     \subfigure[Density Map]{
     \begin{minipage}[b]{0.48\columnwidth} 
        \includegraphics[width=\columnwidth]{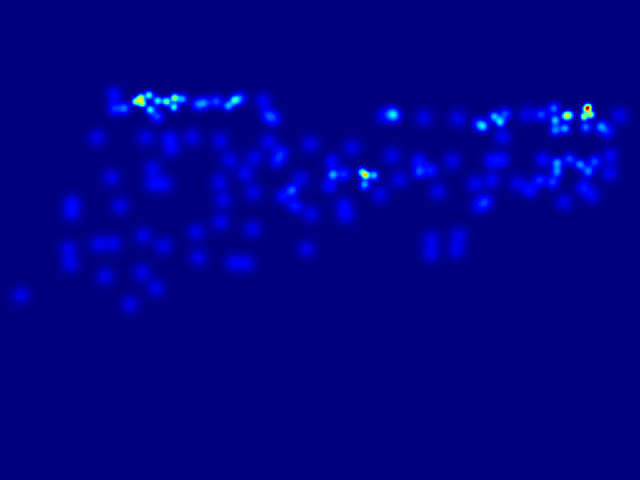}\\
      \includegraphics[width=\columnwidth]{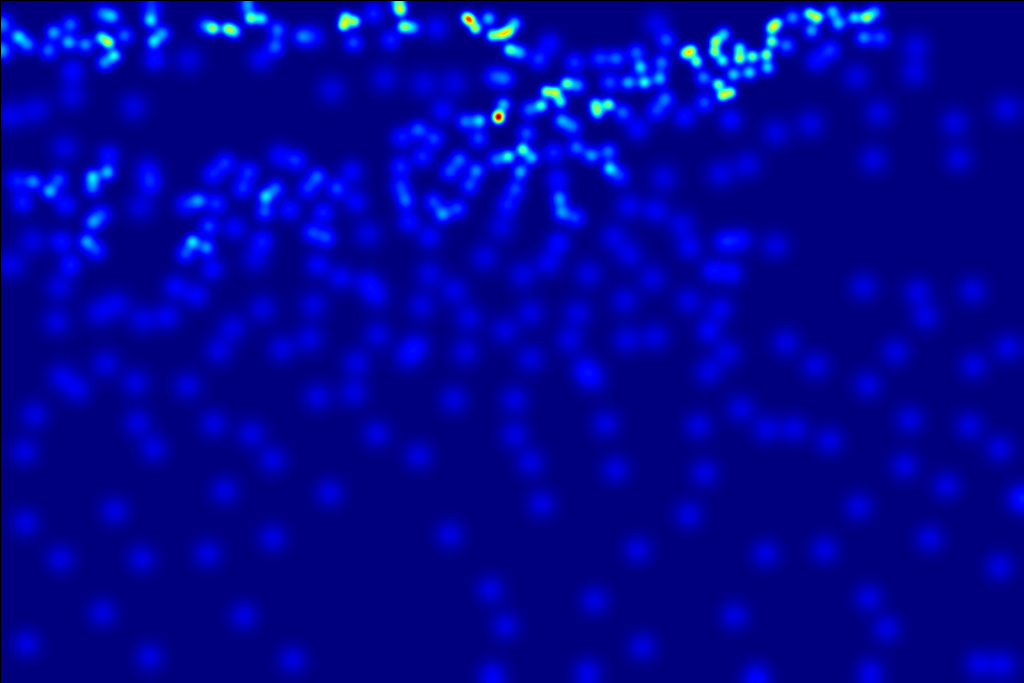}
        \end{minipage}
    }
     \subfigure[Head centers by KMeans]{
     \begin{minipage}[b]{0.48\columnwidth} 
        \includegraphics[width=\columnwidth]{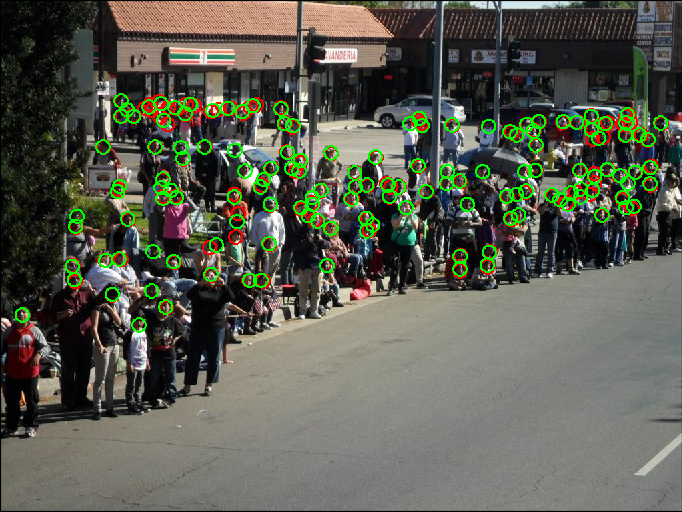}
        \\
      \includegraphics[width=\columnwidth]{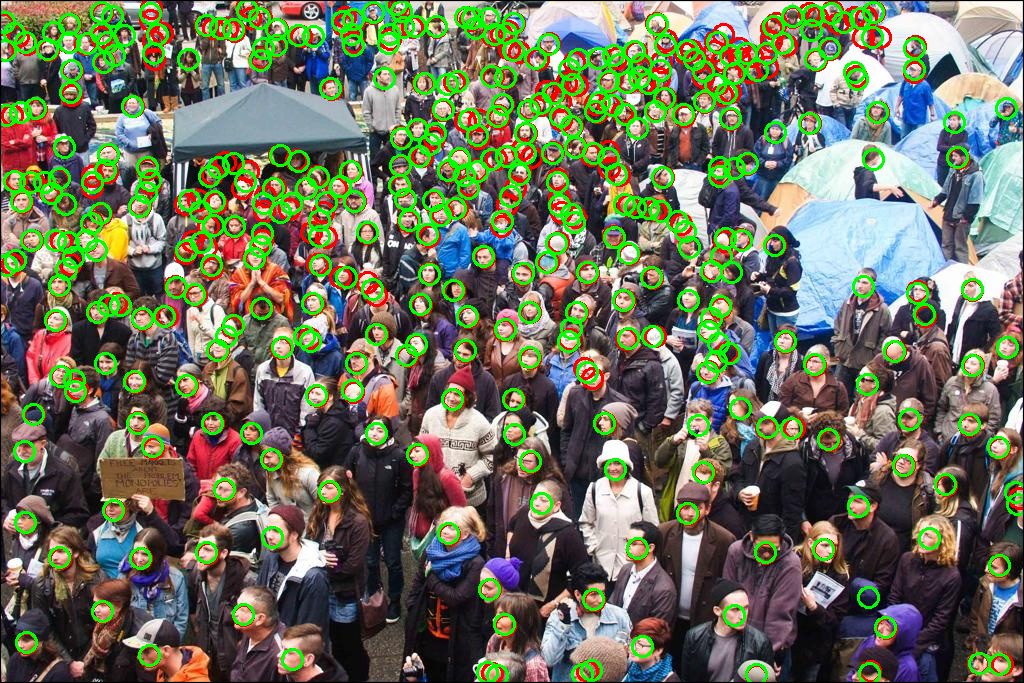}
        \end{minipage}
    }
    \subfigure[Subregions and centers by isolated KMeans]{
     \begin{minipage}[b]{0.48\columnwidth} 
        \includegraphics[width=\columnwidth]{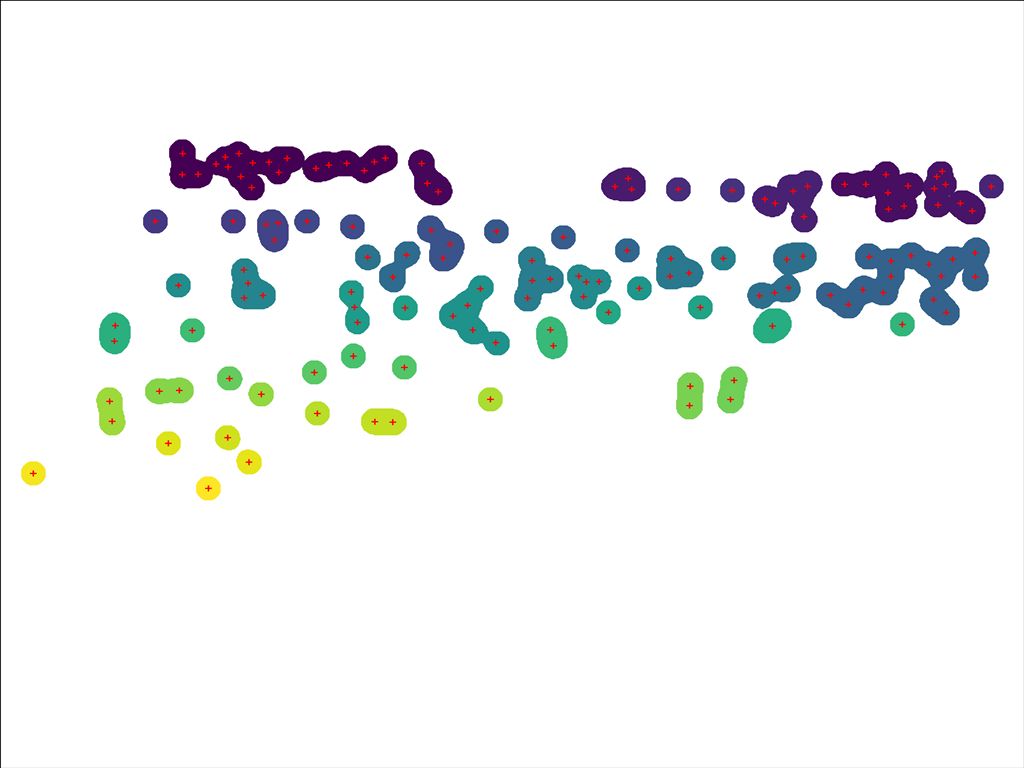}
        \\
      \includegraphics[width=\columnwidth]{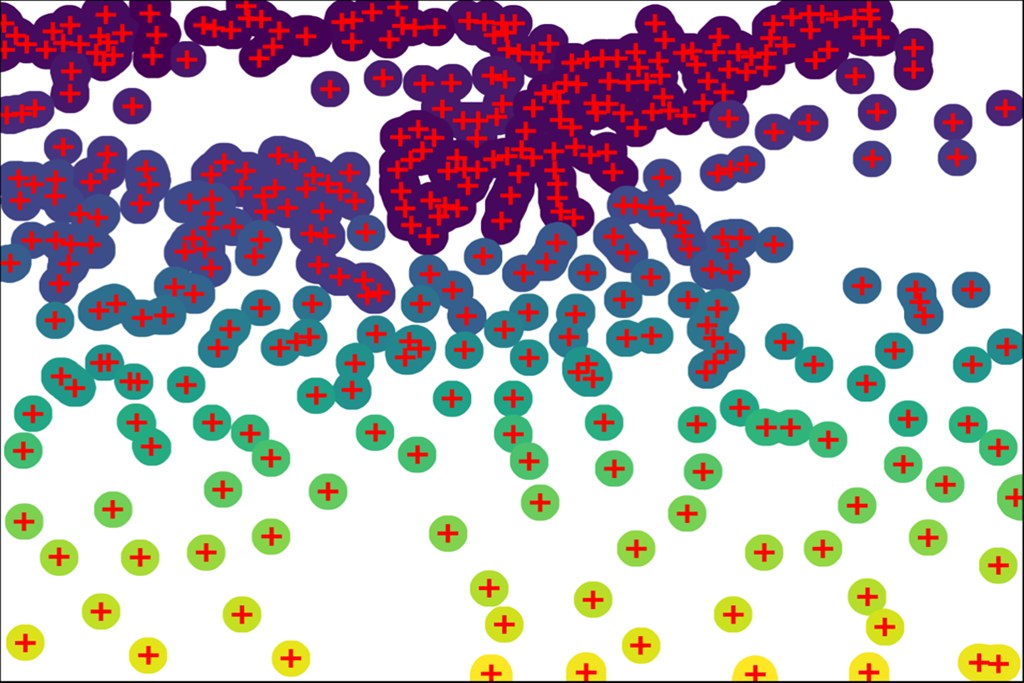}
        \end{minipage}
    }
    \subfigure[Head centers by isolated KMeans]{
     \begin{minipage}[b]{0.48\columnwidth} 
        \includegraphics[width=\columnwidth]{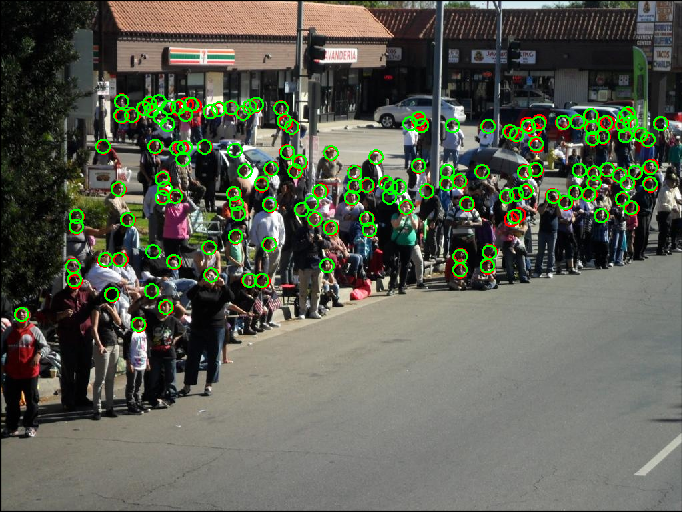}
        \\
      \includegraphics[width=\columnwidth]{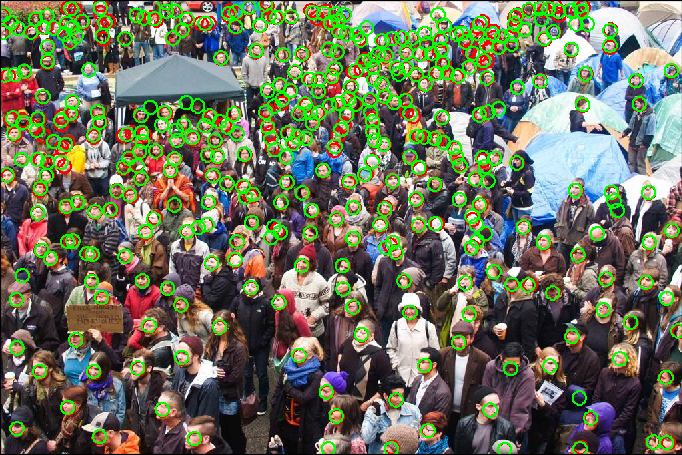}
        \end{minipage}
    }
     \caption{Visualization of the unsupervised localization algorithm. In the second and last column, red circles are the ground truth positions, and green circles are the estimated positions of heads. In the three column, different color means different subregions and '+' means the centers in each subregion.}
     \label{fig:kmeans}
 \end{figure*}

Estimated people count $n$ is calculated by summing all pixel values for a given density map, which means there are about $n$ people in it. Consequently, it is convincing to use $K = n$ as the number of clusters. We use KMeans cluster algorithm as a baseline to locate the heads based on the sampled point set. Typical cluster results are shown in the second column in Figure \ref{fig:kmeans}.

\begin{table}[htbp]
    \centering
    \begin{tabular}{lrrrr}
    \toprule
    Datasets&$\delta$&KMeans&Isolated KMeans\\
    \midrule
    ShanghaiTech PartA&40&83.8\%&\textbf{88.3\%}\\
    &20&75.1\%&\textbf{81.6}\%\\
    &10&36.9\%&\textbf{44.2\%}\\
    \midrule
    ShanghaiTech PartB&40&90.7\%&\textbf{94.3}\%\\
    &20&81.3\%&\textbf{87.0\%}\\
    &10&54.5\%&\textbf{63.9\%}\\
    \midrule
    UCF-QNRF&40&75.1\%&\textbf{78.0\%}\\
    &20&50.7\%&\textbf{54.6\%}\\
    &10&23.9\%&\textbf{26.1\%}\\
    \midrule
    UCF\_CC\_50&40&70.3\%&\textbf{73.9\%}\\
    &20&63.4\%&\textbf{66.8\%}\\
    &10&30.7\%&\textbf{35.9\%}\\
    \bottomrule
    \end{tabular}
    \caption{Localization performance on three crowd benchmarks of KMeans and isolated KMeans in terms of mAP.}
    \label{tab:localization}
\end{table}

Localization precision is calculated in a similar way as the evaluation metric used in MS-COCO \cite{lin2014microsoft}: 1) all centers are ranked according to the number of point in corresponding cluster. 2) from the center with the most points to the one with the least points, we match each center to the nearest person in ground truth. 3) calculate the overlap of the two windows of size $\delta$ centered at the matched points pair. All labelled persons in ground truth can only be matched once. Average Precision(AP) of KMeans-based localization algorithm is reported in table \ref{tab:localization}.

\subsection{Isolated KMeans}

\begin{algorithm}[htbp]
\caption{Isolated KMeans algorithm for locating heads.}
\label{alg:isolated_kmeans}

\begin{algorithmic}[1] %[1] enables line numbers
\STATE Calculate the whole cluster number $K = n$ by summing all the pixel values in the density map $M$.
\STATE Construct the points set $S$ by multiplying the density map $M$ with an expansion factor $k$ and rounding.
\STATE Use DBSCAN to divide $S$ into $N_{sub}$ clusters.
\FOR{sub-point-set $S_{i}$ in cluster 1,..., cluster $N_{sub}$}
\STATE Calculate cluster number $K_{i}$ in this subregion: $K_{i}=\sum_{j\, in\, S_{i}}M_{j}$
\ENDFOR
\STATE Sort the subregions according to $K_{i}$ by ascending order and get cluster 1',..., cluster $N_{sub}^{'}$.
\FOR{sub-point-set $S_{i}^{'}$ in cluster 1',..., cluster $N_{sub}^{'}-1$}
\STATE Use KMeans to find $K_{i}^{'}$ centers in this subregion.
\ENDFOR
\STATE Use KMeans to find $K-\sum_{i=1^{'}}^{N_{sub}^{'}-1}K_{i}$ centers in the last subregion.
\STATE \textbf{return} Positions of $K$ heads.
\end{algorithmic}
\end{algorithm}

Although people count $n$ can match the number of centers $K$ for the whole map in KMeans, we can see that in some regions the cluster number cannot match the sum of these pixel values. In order to make the cluster centers more consistent with the crowd distribution, we divide the whole density map into several subregions firstly, and then use KMeans to locate the heads in each subregion separately. In this way, people count can be matched with the cluster number both globally and locally.

To be specific, we use DBSCAN of $\epsilon=5$ to divide the point set into several clusters. Points in each cluster are regarded as in a subregion. In each subregion, the people number ,i.e. cluster number, is calculated by summing the pixel values in this region and then KMeans can be used to locate the heads. This is illustrated in Algorithm \ref{alg:isolated_kmeans}. AP of isolated KMeans is also calculated in the same way as mentioned before and reported in \ref{tab:localization}. It can be viewed that the localization accuracy has been improved for all $\delta$ parameters. Typical results of divided subregions and their centers and localization results of isolated KMeans are illustrated in Figure \ref{fig:kmeans}.

\section{Conclusion}
In this paper, we collect and evaluate a series of backbones and training tricks in training a neural network for crowd counting. By selecting the best backbone and applying effective training tricks together, we construct an efficient and accurate baseline which improve the MAE and RMSE significantly on three mainly used datasets. We also propose an unsupervised people localization method named isolated KMeans. This clustering algorithm uses the point set constructed from the density maps which eliminates the time of training detection networks, and can also be integrated with any existing counting method.

\bibliographystyle{named}
\bibliography{ijcai20}
\end{document}